\documentclass[smallextended]{svjour3}       % onecolumn (second format)
\usepackage{amsmath}
\usepackage{amssymb}
\usepackage{graphicx}
\usepackage{color}
\begin{document}\sloppy

\title{Predicting Foreground Object Ambiguity and Efficiently Crowdsourcing the Segmentation(s)}
%\title{Foreground Object: Predicting its Ambiguity and Efficiently Crowdsourcing its Segmentation(s)}

%\subtitle{}

\titlerunning{Predicting Foreground Object Ambiguity}        % if too long for running head

\author{Danna Gurari \and 
	Kun He \and
	Bo Xiong \and 
	Jianming Zhang \and 
	Mehrnoosh Sameki \and 
	Suyog Dutt Jain \and 
	Stan Sclaroff \and 
	Margrit Betke \and 
	Kristen Grauman}


%\institute{$^1$ University of Texas at Austin, $^2$ Boston University, \small $^3$ Adobe Research}
\institute{Danna Gurari \at
              University of Texas at Austin, School of Information \\
              1616 Guadalupe St, Austin, TX 78701, USA\\
              \email{\{danna.gurari\}@ischool.utexas.edu}
              \and Bo Xiong, \ Suyog Dutt Jain, \ Kristen Grauman \at
              University of Texas at Austin, Computer Science Department \\
              2317 Speedway, Stop D9500 Austin, TX 78712, USA\\
              \email{\{bxiong,suyog,grauman\}@cs.utexas.edu}
           \and Kun He, \ Jianming Zhang, \ Mehrnoosh Sameki, \ Stan Sclaroff, \ Margrit Betke \at
              Boston University, Computer Science Department \\
              111 Cummington Mall,  \\
              Boston, MA 02215, USA \\
           \email{\{hekun,jmzhang,sameki,sclaroff,betke\}@cs.bu.edu}
}

\date{Received: date / Accepted: date}

\maketitle

%%------------------------------------------------------------------------------------------------------------------
\begin{abstract}
%%------------------------------------------------------------------------------------------------------------------
We propose the ambiguity problem for the foreground object segmentation task and motivate the importance of estimating and accounting for this ambiguity when designing vision systems.  Specifically, we distinguish between images which lead multiple annotators to segment different foreground objects (ambiguous) versus minor inter-annotator differences of the same object.
Taking images from eight widely used datasets, we crowdsource labeling the images as ``ambiguous" or ``not ambiguous"  to segment in order to construct a new dataset we call STATIC.  Using STATIC, we develop a system that automatically predicts which images are ambiguous.  Experiments demonstrate the advantage of our prediction system over existing saliency-based methods on images from vision benchmarks and images taken by blind people who are trying to recognize objects in their environment.  Finally, we introduce a crowdsourcing system to achieve cost savings for collecting the diversity of all valid ``ground truth" foreground object segmentations by collecting extra segmentations only when ambiguity is expected.  Experiments show our system eliminates up to 47\% of human effort compared to existing crowdsourcing methods with no loss in capturing the diversity of ground truths.
\end{abstract}

%%------------------------------------------------------------------------------------------------------------------
%%------------------------------------------------------------------------------------------------------------------

%%------------------------------------------------------------------------------------------------------------------
\section{Introduction}
\begin{figure*}[b]
\centering
\includegraphics[width=1\textwidth]{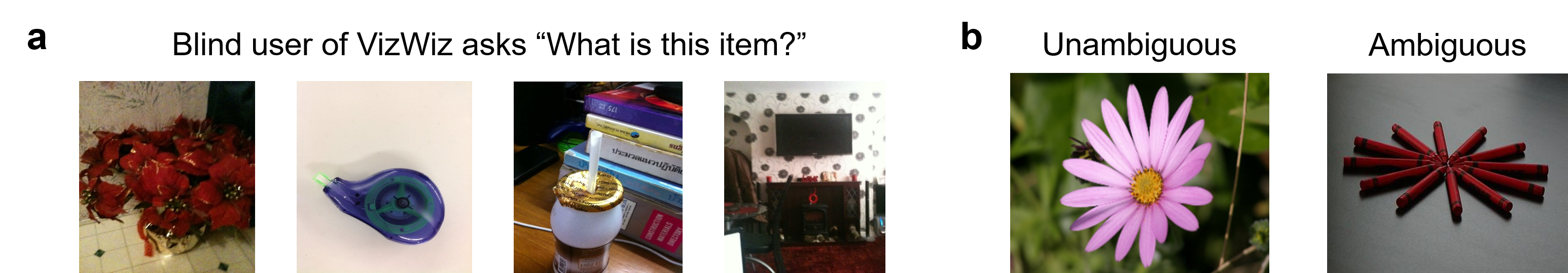}
\caption{Our goal is to build a vision system that can predict from an image whether it is (un)ambiguous where is the foreground object. (\textbf{a}) One benefit of our proposed system is to inform a blind person with high precision if it is ambiguous what object (s)he is trying to recognize with existing mobile phone applications~\cite{TapTapSee,BighamJaJiLiMiMiMiTaWhWhYe10}, empowering him/her to save the 90 seconds typically required with existing systems to learn this from crowd workers~\cite{BighamJaJiLiMiMiMiTaWhWhYe10}.  (\textbf{b}) Images shown are deemed by human judges to have an unambiguous (left) and ambiguous (right) foreground object.  The two images exemplify that objects with similar boundary complexity can differ in whether they are ambiguous.}
\label{fig_motivation}
\end{figure*}

%%------------------------------------------------------------------------------------------------------------------
Finding the most prominent object in an image\footnote{Also, some times referred to as \emph{salient object detection}~\cite{BorjiChJiLi15,ChenMiHuToHu15}} is a critical step for a variety of applications, such as human-robot interaction~\cite{MegerFoLaHeMcSoBaLiLo08}, image retrieval~\cite{ChenMiHuToHu15}, sketch-based image generation~\cite{ChenChTaShHu09}, and assisted recognition for blind people~\cite{LookTelRecognizer,TapTapSee,BradyMoZhWhBi13}.  For example, applications such as VizWiz~\cite{BighamJaJiLiMiMiMiTaWhWhYe10} and TapTapSee~\cite{TapTapSee} enable a blind person to take a picture with a mobile phone and learn ``what is this item?", but these applications depend on the ability to first determine what object a blind person is referring to (\textbf{Figure~\ref{fig_motivation}a}).  In general, a variety of applications rely on first finding the most prominent object in an image \emph{as a human would} (rather than finding all objects~\cite{MartinFoTaMa01} or only finding pre-specified types of objects~\cite{EveringhamGoWiWiZi10,LinMaBeHaPeRaDoZi14,msrc}).

Unfortunately, the aim to build machines that imitate a human's ability to find the most prominent object introduces another problem: how to resolve disagreement that arises when multiple people perceive different foreground objects in the same image.  This problem has spurred an active area of research for methods that combine multiple human annotations in an attempt to recover a single, latent object segmentation as ground truth~\cite{BiancardiJiRe10,CholletiGoBlPoDoSmPr09,WarfieldZoWe04,WelinderPe10}.

In this paper, rather than try to coerce multiple human inputs into a single ground truth, we instead ask, ``Why and when are we observing multiple foreground object segmentations from different annotators?"  We postulate that inconsistent annotations are not only a consequence of difficult tasks and imperfect human annotators, but also a consequence of inherent \emph{ambiguity}.  While psychology research shows humans can perceive foreground object ambiguity~\cite{Perez89,LeopoldWiMaLoBl04}, modern vision systems do not yet account for it.  We aim to fill this gap.  We say \emph{an image is unambiguous if it has a single, non-controversial foreground object of interest}.  
  
%As we will see below, foreground ambiguities are nonetheless common.  
% One extreme example illustrating foreground ambiguity, that is commonly studied in the psychology literature, is Rubin's face-vase illusion---namely, equally valid, yet distinct interpretations of the most prominent object are the vase and two faces~\cite{Perez89}.  As discussed in the literature~\cite{LeopoldWiMaLoBl04}, foreground ambiguities are not as stark as those in Rubin's illusion for natural images, yet are nonetheless common

A central aim of our work is to disentangle the problem of foreground object segmentation \emph{ambiguity} from the problem of foreground object segmentation \emph{difficulty} as the source of human disagreement.  The two problems are only somewhat correlated.  See \textbf{Figure~\ref{fig_motivation}b}.  We observe that two images that are similarly difficult (tedious, complex, etc.) to annotate need not both be ambiguous.  For example, the flower (left image) and the crayons (right image) have similar boundary complexity and exhibit a large ``thing" in the center of the image, yet people deem the flower image unambiguous and the crayon image ambiguous.  The latter may yield segmentations showing any number of the individual crayons or the collection as a whole.  Consequently, human disagreement will likely be far greater when people perceive ambiguity (e.g., crayons) than a single, unambiguous foreground object (e.g., flower).
%\footnote{Each row in \textbf{Fig.~\ref{fig_motivation}} shows one example originating from the MSRA-B, MSRC, BSD, and VizWiz datasets, in that order.}

In light of these considerations, we aim to address two key questions: (1) Given an image, can a machine be trained to predict whether multiple people would identify different foreground object segmentations? and (2) If a machine can automatically predict whether a novel image is ambiguous to people, how might this influence the way we go about obtaining its foreground annotations?  In particular, can we collect annotations that are both economical \emph{and} more complete by knowing when a greater number of segmentations is needed?

To answer these questions, we first introduce a crowd-labeled dataset of nearly 14,000 images, each one annotated as leading to either unambiguous or divergent manual foreground object segmentations.  Included are images from seven existing computer vision benchmarks~\cite{AlpertGaBaBr07,weizmannhorses,EveringhamGoWiWiZi10,GulshanRoCrBlZi10,LiuYuSuWaZhTaSh11,MartinFoTaMa01,msrc} and images taken by blind people who were seeking answers to their daily visual questions with VizWiz~\cite{BighamJaJiLiMiMiMiTaWhWhYe10}.  We then demonstrate the importance of producing datasets with multiple ground truth foreground object segmentations for ambiguous images to avoid biasing algorithms to one interpretation of the truth (\textbf{Sec.~\ref{sec:static}}).  

% We show the promise of a ``shortcut" approach to crowdsource our task, obtaining ambiguity labels \emph{without} requesting multiple hand-drawn segmentations per image 
Next, we leverage the new dataset to develop multiple prediction systems to infer whether an image's foreground object is ambiguous.  Comparing an array of classifiers and features, we report encouraging results for solving the prediction task on both images curated from the web and from blind people (\textbf{Sec.~\ref{sec:predict}}).  Finally, building on these results, we propose a new task of ``redundancy allocation" to capture diversity.  The idea is to exploit our system's ambiguity predictions to decide when multiple human-drawn foreground object annotations are necessary to capture the diversity of opinions, versus when they would likely be redundant.  In this way, we can better spend an annotation budget (\textbf{Sec.~\ref{sec_crowdsourcingSegProblem}}).  Our idea is distinct from prior work that spends an annotation budget to increase confidence in a latent \emph{single} true annotation per image~\cite{SheshadriLease13,WelinderPe10,WhitehillWuBeMoRu09}.  Instead, we spend our budget to efficiently capture the \emph{diversity} of all valid foreground object segmentations for a batch of images.

Our key contributions are the following:

\begin{itemize}
\item Identifying the problem of ambiguity for foreground object segmentation and demonstrating its prevalence for eight diverse datasets.
\item Classification-based approach to identify which images are likely to lead to consistent foreground object segmentation results from multiple humans.
\item System that efficiently captures the diversity of valid foreground object segmentations by soliciting extra manual segmentations only if an image is ambiguous. 
\end{itemize}

%%--------------------------------------------------------------------------------
\section{Related Work}
\label{sec_RelatedWork}
%%--------------------------------------------------------------------------------
\vspace{-0.5em}\paragraph{Defining Foreground Object Segmentation.}
The aim of foreground object segmentation is to produce a binary mask that separates pixels of the most prominent object from the background---also often referred to as salient object detection~\cite{AlpertGaBaBr07,ChenMiHuToHu15,LiuYuSuWaZhTaSh11}.    
Our focus on finding the most prominent object according to human perception is distinct from semantic segmentation where the aim is to segment regions according to a pre-defined set of object categories~\cite{EveringhamGoWiWiZi10,LinMaBeHaPeRaDoZi14,msrc}.  Our task is also distinct from natural scene segmentation where bottom-up methods are employed to segment an image into any number of regions~\cite{MartinFoTaMa01}.  Foreground object segmentation also differs from edge detection methods, e.g.,~\cite{DollarZi15}.  To our knowledge, we are the first to propose the problem of deciding \emph{whether a single, unambiguous salient object exists} in an image.  Knowing whether an image shows a single, \emph{unambiguous} foreground object is critical for the success of many applications, such as human-robot interaction, image retrieval, sketch-based image generation, and assisted recognition for blind people~\cite{LookTelRecognizer,TapTapSee,BorjiChJiLi15,ChenMiHuToHu15}. 
    
\vspace{-0.5em}\paragraph{Predicting Ambiguity.}    
Other work in computer vision explores ambiguity in relationship to language.  This includes whether images lead to more or less ``specific" text descriptions~\cite{JasPa15} and whether visual attributes permit multiple interpretations~\cite{KovashkaPaGr14}.  While these prior works predict image ambiguity related to language, our work predicts image ambiguity for foreground object segmentation.    
%Our work relates to computer vision methods that explore ambiguity, including whether images lead to more or less ``specific" text descriptions~\cite{JasPa15}, visual attributes permit multiple interpretations~\cite{KovashkaPaGr14}, or visual questions permit multiple answers~\cite{GurariGr17}.  Unlike prior work, our work predicts ambiguity for foreground object segmentation.    

\vspace{-0.5em}\paragraph{Predicting Image Segmentation Difficulty.}
A related, yet distinct problem in modern computer vision literature is predicting segmentation \emph{difficulty}, where difficulty is commonly defined by the extent to which algorithms can produce segmentations similar to the ground truth for a given image~\cite{JainGr13,KohlbergerSiAlBaGr12,LiuXiPuSh11} or the time a person takes to segment an image~\cite{VijayanarasimhanGr11}.  However, what can be deemed a successful method for predicting segmentation difficulty may be a ``moving target", given the development of better algorithms and easier-to-use segmentation annotation systems.  In contrast, since we aim to capture human-perceived ambiguity, our method to estimate ambiguity directly measures an intrinsic property about an image and so leads to a static ``yes" or ``no" outcome (rather than an evolving ranking based on the chosen algorithm or annotation system).

\vspace{-0.5em}\paragraph{Establishing Ground Truth.}
The status quo when creating ground truth with crowdsourcing is to collect redundant annotations.  This is because discrepancies in human-provided annotations are anticipated, whether due to crowd worker skill or bias~\cite{SheshadriLease13,WelinderPe10,WhitehillWuBeMoRu09}; hence, the goal in prior work is to discover the \emph{single latent ground truth} for each example in spite of those discrepancies.  Consequently, many methods intelligently sample and fuse labels from multiple workers in an attempt to produce a final high-quality annotation~\cite{WelinderBrBePe10,WelinderPe10}.  In contrast, we address discrepancies that stem from \emph{ambiguity}, meaning that there does not exist a single latent ground truth for each image.  As such, our goal is not to gather enough annotations to wipe away annotator differences~\cite{SheshadriLease13,WelinderPe10,WhitehillWuBeMoRu09}; rather, it is to collect (just) enough annotations to capture annotator differences.  Our results demonstrate this important distinction.  

Our work more closely relates to the pioneering segmentation collection work by Martin et al.~\cite{MartinFoTaMa01}, who collected multiple segmentations of natural scenes from independent annotators, motivated by the belief that segmentation tasks can afford multiple correct answers.  Whereas Martin et al. gathered a fixed number of annotations for each image from known in-house annotators to provide a soft ground truth for image contours, we show both how to predict which images offer multiple interpretations for the foreground object segmentation problem and how to more economically collect redundant annotations from an anonymous on-line crowd.  

\vspace{-0.5em}\paragraph{Crowdsourcing Object Segmentation Collection.}
Numerous systems already collect object segmentations from online crowds, including LabelMe~\cite{RussellToMuFr08} and the MSCOCO crowdsourcing pipeline~\cite{LinMaBeHaPeRaDoZi14}.  These systems instruct the worker to segment as many objects as (s)he chooses~\cite{RussellToMuFr08} or as many instances of a given object category (s)he observes~\cite{LinMaBeHaPeRaDoZi14}.  In both cases, the aim is to efficiently segment and name \emph{all relevant objects} in a given multi-object scene image.  In contrast, the goal of our system is to efficiently capture the diversity of human opinions on the \emph{single, most salient object} for a given image.  Consequently, as commonly done in human computation systems~\cite{LiuYuSuWaZhTaSh11,MartinFoTaMa01}, we collect annotations from multiple, independent annotators to avoid biasing workers.  However, in contrast to these human computation systems, we automatically predict how many independent annotators to recruit to efficiently complete the task.

\vspace{-0.5em}\paragraph{Blind Photography.}
Numerous systems have been proposed to assist blind people to take a high quality picture of an object with a mobile phone camera~\cite{TapTapSee,BighamJaJiLiMiMiMiTaWhWhYe10,JayantJiWhBi11,VazquezSt14,ZhongGaBi13}.  Unfortunately, such systems assume a user can localize the desired object and only help the user to improve the image focus~\cite{TapTapSee}, lighting~\cite{BighamJaJiLiMiMiMiTaWhWhYe10}, or composition~\cite{JayantJiWhBi11,VazquezSt14,ZhongGaBi13}.  Unlike prior work, we do not assume a user can localize the object of interest.  Rather, we propose a method that can be employed to automatically alert a blind user whether an image shows a single, unambiguous object.  We demonstrate the predictive advantage of our system for this task over relying on saliency-based methods~\cite{FengWeTaZhSu11,ZhangMaSaScBeLiShPrMe15}.   
   
%%--------------------------------------------------------------------------------
\section{STATIC - When Is There a Single Truth?}\label{sec:static}
%%--------------------------------------------------------------------------------
In this section, we first present our crowdsourcing dataset collection process (\textbf{Sec.~\ref{sec_ambiguousImageLabeling}}) to label segmentation ambiguity on images from multiple existing benchmarks.  Then, we examine how the labels compare to ambiguity labels derived using multiple human-drawn segmentations (\textbf{Sec.~\ref{sec_communityConsistencyStudy}}). Finally, we analyze how foreground ambiguity, as perceived by humans, influences the evaluation of segmentation algorithms (\textbf{Sec.~\ref{sec_labelingImgDifficultyForComputers}}).  \textbf{Sections \ref{sec:predict}} and \textbf{\ref{sec_crowdsourcingSegProblem}} will introduce our ideas for a machine to predict for a novel image whether it has an ambiguous foreground object and then to efficiently collect the diversity of all valid foreground object segmentations for a batch of images.

\subsection{Judging When an Image is Ambiguous}
\label{sec_ambiguousImageLabeling}

\begin{figure*}[b!] 
\centering
\includegraphics[width=1\textwidth]{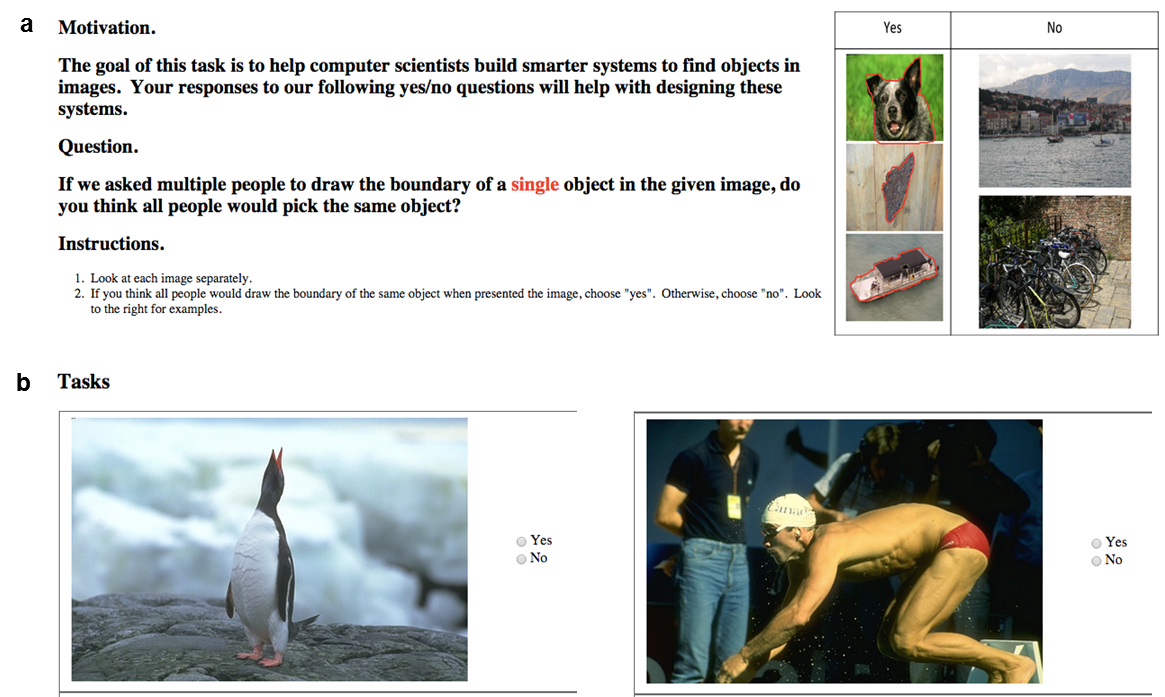}
\caption{(\textbf{a}) Instructions and (\textbf{b}) user interface shown to crowd workers for the voting task to judge whether an image shows a single, unambiguous foreground object.}
\label{fig_crowdVotingInstructionsAndUI}
\end{figure*}   

\vspace{-0.5em}\paragraph{Crowdsourcing Strategy.}
The traditional method to identify human (dis)agreement is to solicit multiple people to annotate the same image and then analyze the consistency between the multiple object segmentations~\cite{AlpertGaBaBr07,LiuYuSuWaZhTaSh11,MartinFoTaMa01}.  However, evaluating if multiple people will agree on a single, unambiguous foreground object based on multiple annotations is less direct than simply asking them what they perceive.  Moreover, collecting multiple object segmentation masks is labor-intensive and costly.  We instead explicitly ask an annotator to judge, for a given image, whether (s)he thinks the image segmentation task would lead to a diversity of foreground objects from multiple annotators.  Our use of less costly, human judgments over evaluating annotation results aligns with existing crowdsourcing work~\cite{Gilbert14,ShawHoCh11}.  We call our approach and our dataset a {\bf S}egmentation {\bf T}est for {\bf A}mbiguous {\bf T}ruth {\bf I}nferred for the {\bf C}rowd  (\textit{STATIC}).   Each image receives a binary label indicating whether it has a single, unambiguous object segmentation truth, based on human opinion.  

We collect image labels from on-line crowd workers on Amazon Mechanical Turk (AMT).  We designed our Human Intelligence Task (i.e., HIT) with instructions followed by the voting task (\textbf{Fig.~\ref{fig_crowdVotingInstructionsAndUI}}).  We include five images per HIT.  For the voting task, we ask workers the following question: ``If we asked multiple people to draw the boundary of a single object in the given image, do you think all people would pick the same object?"  We intentionally specify criteria that aligns with the generic object segmentation task.  A crowd worker casts a vote by selecting one of two radio buttons to the right of each image to indicate ``Yes" or ``No."  To minimize concerns about worker skill, we limit our pool of workers to those who previously completed at least 100 tasks and received at least a 92\% approval rating.  To address concerns about malicious crowd workers, we collect five predictions per image and then assign the majority vote label.  We pay workers \$0.02 to complete each HIT.  

Our \textit{STATIC} labeling approach is advantageous not only because it offers cost and time savings, but also because (1) it disentangles the segmentation ambiguity problem from the many other factors that can lead to disagreement; for example, annotator training/skill, segmentation difficulty and (2) it avoids potential biases that may arise when soliciting a small number of humans to segment objects (e.g., \textbf{Fig.~\ref{fig_diversityTypes}}).  For example, workers may annotate what is easiest to minimize segmentation effort.  Our analysis in the next section (\textbf{Sec.~\ref{sec_communityConsistencyStudy}}) shows our shortcut of explicitly asking workers for the ambiguity label can successfully produce high quality labels.  

\vspace{-0.5em}\paragraph{Dataset Construction.}
We built \emph{STATIC} from eight publicly-available datasets.  We include seven widely-studied computer vision segmentation benchmarks~\cite{AlpertGaBaBr07,EveringhamGoWiWiZi10,GulshanRoCrBlZi10,LiuYuSuWaZhTaSh11,MartinFoTaMa01,msrc,weizmannhorses} in order to enrich them with ground truth about image ambiguity.  We also include a dataset of images taken by blind people with mobile phone cameras via VizWiz~\cite{BighamJaJiLiMiMiMiTaWhWhYe10} in order to study foreground object ambiguity in the context of an important practical problem of assisting blind people to take a picture of an object.  

\emph{STATIC} includes three computer vision benchmarks designed to contain images with a single object of interest~\cite{AlpertGaBaBr07,LiuYuSuWaZhTaSh11,weizmannhorses}.  These benchmarks were created to evaluate foreground object segmentation algorithms~\cite{AlpertGaBaBr07,weizmannhorses} and salient object detectors~\cite{LiuYuSuWaZhTaSh11}.  In particular, Weizmann~\cite{AlpertGaBaBr07} contains grayscale images showing a variety of everyday objects, and Weizmann Horses~\cite{weizmannhorses} and MSRA-B~\cite{LiuYuSuWaZhTaSh11} consist of RGB images showing horses and a variety of everyday images respectively.  As we will see below, though a single prominent object is expected in these datasets, that is not always how each image is perceived.

\emph{STATIC} also includes four computer vision benchmarks that were created to evaluate algorithms for natural scene segmentation~\cite{MartinFoTaMa01}, interactive image segmentation~\cite{GulshanRoCrBlZi10}, and semantic segmentation~\cite{EveringhamGoWiWiZi10,msrc}.  A priori, we expect these datasets to offer greater ambiguity since they are not designed to contain a single object of interest.  All datasets contain RGB images, with Berkeley Segmentation Dataset~\cite{MartinFoTaMa01} (BSD) showing natural scenes, Interactive Image Segmentation~\cite{GulshanRoCrBlZi10} (IIS) showing a variety of everyday objects, and MSRC~\cite{msrc} and VOC2012~\cite{EveringhamGoWiWiZi10} showing everyday scenes of 23 and 20 object classes respectively.  Again, as we will see below, even though many images have multiple objects, some have an unambiguous foreground object and some do not.  

Finally, \emph{STATIC} includes 4,163 randomly selected images from the VizWiz dataset~\cite{BighamJaJiLiMiMiMiTaWhWhYe10} that were taken by blind people with mobile phone cameras to learn answers to their visual questions\footnote{We excluded all images for which the majority of three crowd workers indicated the answer to their visual question could be recognized by text in the image.}.  These images often are poor quality due to poor framing, poor lighting, and motion blur.  Nonetheless, these images capture a real world scenario where individuals are typically trying to recognize an object in their environment.  Specifically, approximately 65\% of the VizWiz images were captured because a blind person wanted to either identify an object (e.g., ``What is this item?") or have an object described (e.g., ``What color is this shirt?")~\cite{BradyMoZhWhBi13}.  In other words, the blind photographer typically intended to capture a single, most prominent object.

In total, \emph{STATIC} includes 13,746 images; \textbf{Table~\ref{table_crowdPerceptionStudy}} shows the breakdown.  The resulting collection includes images showing a single object, multiple objects in possibly complex scenes, or no object of interest in blurry or poorly lit images.  As will be shown in the next section, this diversity of image content is valuable for training classifiers to accurately decide if an image shows a single, unambiguous object.    

% We publicly share this new dataset ({\tt http://Anonymous}).    

% While segmenting the foreground object in an image is distinct from the tasks of semantic segmentation~\cite{}, interactive image segmentation~\cite{}, and edge detection~\cite{}, we will show that 49\% to 58\% of images in community-shared benchmarks for these related problems are foreground object segmentation problems.  

%- We intentionally built STATIC from 7 datasets that comprise scene-centric and object-centric datasets (L213-42). Like R6, we expected more ambiguity in scene datasets (BSD, MSRC, VOC, IIS). Yet, the data reveals 49% to 58% of those images are judged to have a single, unambiguous object (Tbl 1).

\vspace{-0.5em}\paragraph{Dataset Characterization.}
We use the crowdsourcing strategy above to obtain human judgments about foreground object segmentation ambiguity.  \textbf{Table~\ref{table_crowdPerceptionStudy}} summarizes the results.  As observed, even benchmarks explicitly designed for foreground object segmentation (top three rows) have ambiguity for 5\% to 25\% of images.  On the flip side, it is interesting to note that datasets \emph{not} intentionally built for foreground object segmentation have 36\% to 58\% of images showing a single, unambiguous object.  Our findings highlight that, in a wide range of datasets, some images have ``well-defined" foreground object segmentation truths while others lead to a diversity of viable interpretations.  

\begin{table}[h] \small
        \centering
            \caption{Percentage of images in eight datasets that humans judge to \emph{not} show a single, unambiguous object.  Note, only the first three datasets were explicitly created with the intention of containing a single foreground object per image.  As observed, benchmarks explicitly designed for both foreground object segmentation (top three rows) and complex scenes contain a mix of ambiguous and unambiguous images.} 
            \vspace{0.25em}
    \begin{tabular}{| l | c | c | c | c |}
   \hline
       & {\bf \# Images} & {\bf \# Workers} & {\bf \% Ambiguous Images}   \\  \hline   \hline
      \textbf{Horses~\cite{weizmannhorses}} & 328 & 33 & 5\% (ambiguity unexpected) \\ \hline
      \textbf{Weizmann~\cite{AlpertGaBaBr07}} & 100 & 25 & 19\% (ambiguity unexpected)\\ \hline
      \textbf{MSRA-B~\cite{LiuYuSuWaZhTaSh11}} & 5,000 & 128 & 25\% (ambiguity unexpected) \\ \hline \hline
      \textbf{IIS~\cite{GulshanRoCrBlZi10}} & 151 &  10 & 42\%  \\ \hline
      \textbf{VOC2012~\cite{EveringhamGoWiWiZi10}} & 2,913 & 97 & 43\%  \\ \hline
      \textbf{MSRC~\cite{msrc}} & 591 & 47 & 48\%   \\ \hline
      \textbf{BSD~\cite{MartinFoTaMa01}} & 500 &  25 & 51\% \\ \hline
      \textbf{VizWiz~\cite{BighamJaJiLiMiMiMiTaWhWhYe10}} & 4,163 & 25 & 64\% \\ \hline
    \end{tabular}
    \label{table_crowdPerceptionStudy}
\end{table}

\subsection{Labels: Direct Ambiguity Judgment versus Redundant Segmentations}
\label{sec_communityConsistencyStudy}
We next investigate an important question of whether human judgments about ambiguity, as collected above, correspond to the judgments one would obtain with today's status quo approach of collecting multiple segmentations~\cite{AlpertGaBaBr07,LiuYuSuWaZhTaSh11,MartinFoTaMa01}.  In what follows, we term the crowd workers who declared the images as (un)ambiguous as \emph{judgers}, and the annotators who manually drew segmentations as \emph{drawers}.

We perform our comparison on the Weizmann benchmark~\cite{AlpertGaBaBr07}, since it includes three human-drawn segmentations per image for single object images.  We use the hand-drawn segmentations to produce a ground truth hypothesis parallel to the one created by the judgers.  Namely, we label an image as ambiguous if any drawer segments more than one object or if any pair of drawers segment the single foreground object differently (i.e., less than 50\% intersection-over-union overlap). %, as also done in prior work~\cite{RussellToMuFr08}).

\begin{figure}[b!]
\centering
\includegraphics[width=1\textwidth]{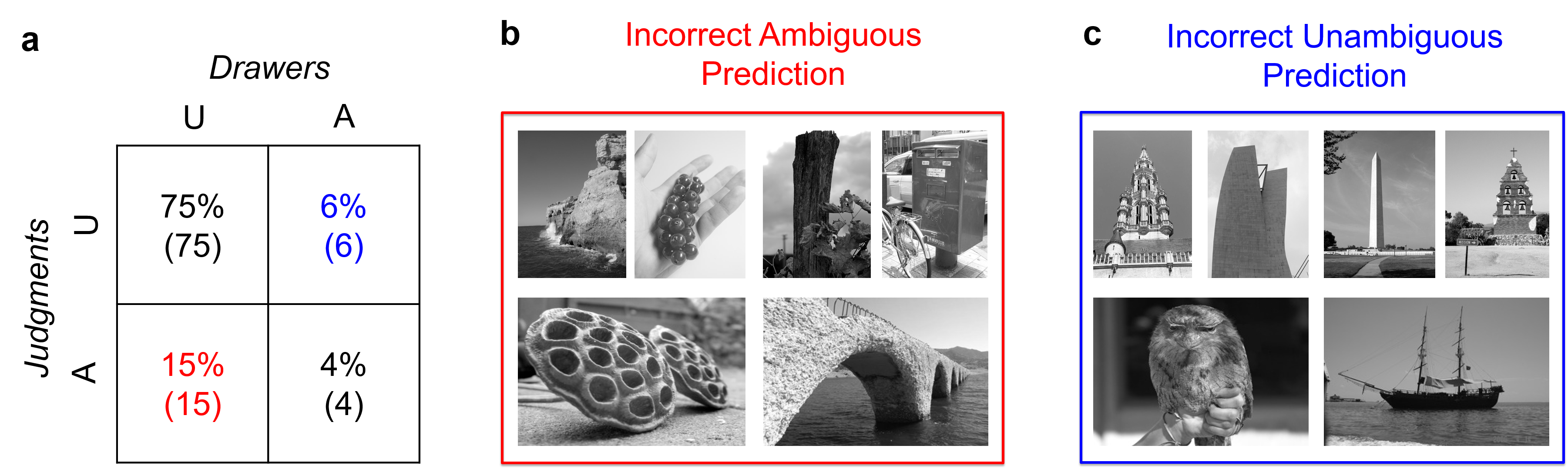}
\caption{Analyzing label agreement between judgers and drawers.  (\textbf{a}) Percentage of Weizmann images deemed ambiguous (A) or unambiguous (U) by either labeling mode.  (\textbf{b}, \textbf{c}) Images illustrating why judgers and drawers may disagree.  (\textbf{b}) shows images that judgers find ambiguous but drawers do not.  (\textbf{c}) shows images that judgers find unambiguous but drawers do not.}
\label{fig_crowdPredVsExpertConfusionMatrix}
\end{figure}

\textbf{Figure~\ref{fig_crowdPredVsExpertConfusionMatrix}(a)} shows the consistency of the two parallel labels.  The matrix (left) breaks down the fraction of images receiving each label (unambiguous (U) or ambiguous (A)) by each party.  Our findings demonstrate that our proposed labeling approach matches labels produced by the status quo approach for 79\% of the images.  Moreover, our approach achieves these high quality labels while reducing the status quo human annotation effort by over a factor of 10 (i.e., 4.7 seconds per judgement versus 50 seconds per drawing~\cite{GurariSaBe16}).  %Our findings complement existing work~\cite{Gilbert14,ShawHoCh11} by similarly demonstrating that directly asking whether a crowd will agree on a single answer can yield high quality results.  

% Overall, for 79\% of the images, we see that the judger labels in \textit{STATIC} agree with the drawer labels.  

%We will mention important future work includes exploring other labelling methods (e.g., R1's suggestion)

%While we will mention important future work includes exploring other labelling methods (e.g., R1's suggestion), our work offers convincing evidence our proposed direction of estimating foreground object ambiguity is worth pursuing further. \vspace{0.2em}

\textbf{Figure~\ref{fig_crowdPredVsExpertConfusionMatrix}(b,c)} show cases where the judger and drawer labels disagree.  Interestingly, the primary reason for label disagreement is because judgers predict drawer disagreement too often.  We attribute the judgers' overzealous labeling of ambiguity to judgers identifying more regularity of known causes of drawer disagreement.  For example, drawers commonly disagree by segmenting at \emph{different granularity levels for the same object}; e.g., while drawers disagreed whether to include the strings on the ship (\textbf{Figure~\ref{fig_crowdPredVsExpertConfusionMatrix}c}; bottom right corner), they did not disagree whether to segment one seed pod or both pods (\textbf{Figure~\ref{fig_crowdPredVsExpertConfusionMatrix}b}; bottom left corner).  In addition, drawers also commonly disagree by segmenting \emph{different primary objects}; e.g, while drawers disagreed whether to segment the building or traffic sign (\textbf{Figure~\ref{fig_crowdPredVsExpertConfusionMatrix}c}; top right corner), they did not disagree whether one would segment the water, cliff, or sky (\textbf{Figure~\ref{fig_crowdPredVsExpertConfusionMatrix}b}; top left corner).  As exemplified in \textbf{Figure~\ref{fig_crowdPredVsExpertConfusionMatrix}}, the crowd judgers may be more effective in detecting plausible ambiguity than what can be revealed by a small sample size (e.g., three drawers).  We further explore this issue of the appropriate sample size to detect ambiguity in \textbf{Section~\ref{sec_crowdsourcingSegProblem}}.

%%--------------------------------------------------------------------------------
\subsection{Impact of Ambiguity on Evaluation of Segmentation Algorithms}
\label{sec_ambiguityAndSegEval}
%%--------------------------------------------------------------------------------
\label{sec_labelingImgDifficultyForComputers}
We finally investigate how foreground object ambiguity may impact how we judge the performance of algorithms that segment foreground objects.

We conduct the study on the Weizmann~\cite{AlpertGaBaBr07} dataset, which has a single foreground object ``ground truth" per image.  We evaluate the following five commonly-employed algorithms against the ground truth using the intersection-over-union measure: Grab Cut~\cite{RotherKoBl04}, level set methods~\cite{CasellesKiSa97,ShiKa08}, an object region proposal method~\cite{ArbelaezPoBaMaMa14}, and a salient object detection method~\cite{LiuYuSuWaZhTaSh11}.  

\begin{figure}[h!]
\centering
\includegraphics[width=1\textwidth]{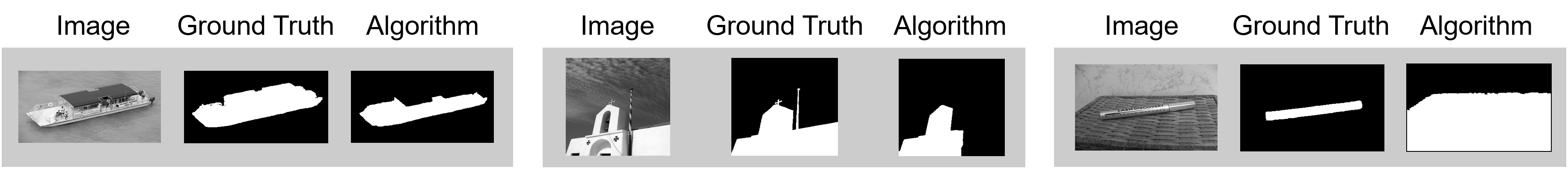}
\caption{Shown are examples for which algorithms disagree with the benchmark ``ground truth" for ambiguous labeled images.  As shown, by not including all valid segmentations as ground truth, we can miss out on learning whether our algorithms are succeeding according to any valid interpretation.}
\label{fig_humanVsComputer}
\end{figure}

\textbf{Figure~\ref{fig_humanVsComputer}} illustrates how a benchmark with a \emph{single ground truth} leads us to judge images as difficult for an algorithm to segment when the algorithm in fact produces a distinct, valid interpretation for how to segment an ambiguous image.  Consistent with this finding, we find that the overall top-performing method from the five algorithms switches from \cite{CasellesKiSa97} to \cite{ArbelaezPoBaMaMa14} when we exclude the ambiguous images from evaluation (19 of the images are labeled as ambiguous).  Our findings demonstrate a problem with evaluating algorithm results against a \emph{single ground truth}.  The use of the phrase ``ground truth" in existing benchmarks may lead our community to miss out on learning whether our algorithms are succeeding---according to any of the viable interpretations---for a significant portion of our benchmark images!  
%While we found that the algorithms are labeled as failing for 63\% of the ambiguous-labeled images (12 out of 19 ambiguous-labeled images), \emph{the very nature of ambiguous foreground object means that algorithms are not necessarily failing; rather, they are (possibly) firing on alternate plausible regions} (e.g., \textbf{Fig.~\ref{fig_crowdPredVsExpertConfusionMatrix}d}).  Our use of the phrase ``ground truth" in existing benchmarks may lead our community to miss out on learning whether our algorithms are succeeding---according to any of the viable interpretations---for a significant portion of our benchmark images!  Accordingly, we found that the top-performing method from the five algorithms switches from \cite{CasellesKiSa97} to \cite{ArbelaezPoBaMaMa14} when excluding the ambiguous images from evaluation.  In light of these concerns, \textit{STATIC} offers a finer-grained ground truth.

% e.g., permit multiple ground truths for \emph{ambiguous} images to avoid incorrectly discounting them a \emph{difficult} for an algorithm

% In existing benchmarks, algorithms may be perceived as failing when they are actually capturing a distinct viable interpretation of how to segment an image.  

%%--------------------------------------------------------------------------------
\section{Predicting Ambiguous Images}\label{sec:predict}
%%--------------------------------------------------------------------------------
Having established a dataset of images annotated for their human-perceived ambiguity, we now turn to the question of whether ambiguity is machine learnable.  We pose the task as a binary classification problem: given a novel image, can we correctly classify it as \emph{ambiguous} or \emph{unambiguous} using only the image content?  To our knowledge, no prior work has directly addressed the problem of predicting foreground object ambiguity.

\vspace{-0.5em}\paragraph{Classifiers and Features.} 
We benchmark a total of nine classifiers.  We include two related saliency methods.  We also train six Support Vector Machine (SVM) classifiers based on both traditional global image features and deep Convolutional Neural Network (CNN) features.  Finally, inspired by the recent success of CNNs for image classification, we propose a \textit{STATIC} fine-tuned CNN classifier.   

The two existing saliency methods we benchmark are the salient object detector of Feng et al.~\cite{FengWeTaZhSu11} and the salient object subitizing (SOS) method of Zhang et al.~\cite{ZhangMaSaScBeLiShPrMe15}.  Intuitively, both should be relevant to our prediction task, since salient object strength and the number of detected salient objects should correlate with (un)ambiguity.  Feng et al.'s system~\cite{FengWeTaZhSu11} outputs a ranked list of detection subwindows.  We improve its results by a refined non-maximum suppression stage, using an aggressive non-maximum suppression threshold of 0.1 to suppress overlapping detections.  When a single window is returned, we use its confidence as the unambiguity score.  Otherwise, we take the difference in scores of the best and second-best detections based on the intuition that a dominant salient object should ``stand out" over other areas in the same image.  The SOS method~\cite{ZhangMaSaScBeLiShPrMe15} fine-tunes the VGG16 CNN~\cite{SimonyanZi14} to produce a probability that the image contains (0, 1, 2, 3, or 4+) salient objects.  We use the probability it returns for 1 object as the output.  

%We use the probability it returns for 1 object as the output.  This is a strong baseline, as the authors of \cite{ZhangMaSaScBeLiShPrMe15} report over 90\% AP in predicting images with no salient object and with a single salient object on their dataset.     
%This is a strong baseline, as the authors of \cite{ZhangMaSaScBeLiShPrMe15} report over 90\% AP in predicting images with no salient object and with a single salient object on their dataset.  An image is deemed unambiguous if SOS predicts there is one salient object in the image with probability greater than a given threshold $X$.

We also test six Support Vector Machine (SVM) classifiers.  Three of the classifiers are trained using off-the-shelf gradient-based, global image features: GIST~\cite{TorralbaMuFrRu03}, HOG~\cite{DalalTr05}, and IFV~\cite{PerronninSaMe10}.  The other three are trained using the 4096-dimensional output from the last fully connected layer of three Convolutional Neural Networks (CNN): AlexNet~\cite{KrizhevskySuHi12}, VGG16~\cite{SimonyanZi14} and SOS~\cite{ZhangMaSaScBeLiShPrMe15}.  For each of the six SVM-based classifiers, we reduce the dimensionality of the feature (GIST, HOG, IFV, AlexNet, VGG16, SOS) to 100 using PCA before applying the SVM classifier.  We use degree 3 polynomial kernels, and apply 5-fold cross validation to choose the SVM hyper-parameters.

Finally, we propose a \textit{STATIC} fine-tuned CNN classifier.  We fine-tune the subitizing (SOS) network to target a binary classification loss on the labeled \textit{STATIC} training images~\cite{caffe}.  We set the starting learning rate to a moderate 0.0001 and fine-tune for 20 epochs.  The subitizing features are attractive for the task at hand since intuitively an unambiguous image might be estimated to have a single salient object (recall that SOS yields a probability that the image contains one salient object).  

To recap, we benchmark nine classification pipelines, including two existing baseline models:
%\vspace{-0.4em}
\begin{description}
\setlength\itemsep{0em}
\item [\texttt{CNN-FT}]: CNN classifier fine-tuned on STATIC.
\item [\texttt{SVM-GIST}]: SVM on GIST.
\item [\texttt{SVM-HOG}]: SVM on HOG.
\item [\texttt{SVM-IFV}]: SVM on IFV.
\item [\texttt{SVM-AlexNet}]: SVM on AlexNet CNN features.
\item [\texttt{SVM-VGG16}]: SVM on VGG16 CNN features.
\item [\texttt{SVM-SOS}]: SVM on the SOS fine-tuned CNN features.
\item [\texttt{Feng et al.}~\cite{FengWeTaZhSu11}]: a salient object detector.
\item [\texttt{SOS}~\cite{ZhangMaSaScBeLiShPrMe15}]: a salient object subitizing (counting) method. 
\end{description}

\vspace{-0.5em}\paragraph{Datasets.} We evaluate all classifiers on both the images coming from the 1) seven computer vision benchmarks and 2) VizWiz dataset.  In doing so, we aim to learn the value of these classification systems both for high-quality, curated images from the web as well as unknown quality images collected from blind photographers with mobile phone cameras.  For both sets of images, we apply a random 80/20 train/test split.  We train each of our first seven classifiers in the list using the same training data.  We use the remaining two methods as is.  At testing, all nine models produce a probability / confidence output per image that we use to evaluate against \textit{STATIC} ambiguity ground truth for generating precision-recall curves. 

\begin{figure}[t!]
\centering
\includegraphics[width=1\textwidth]{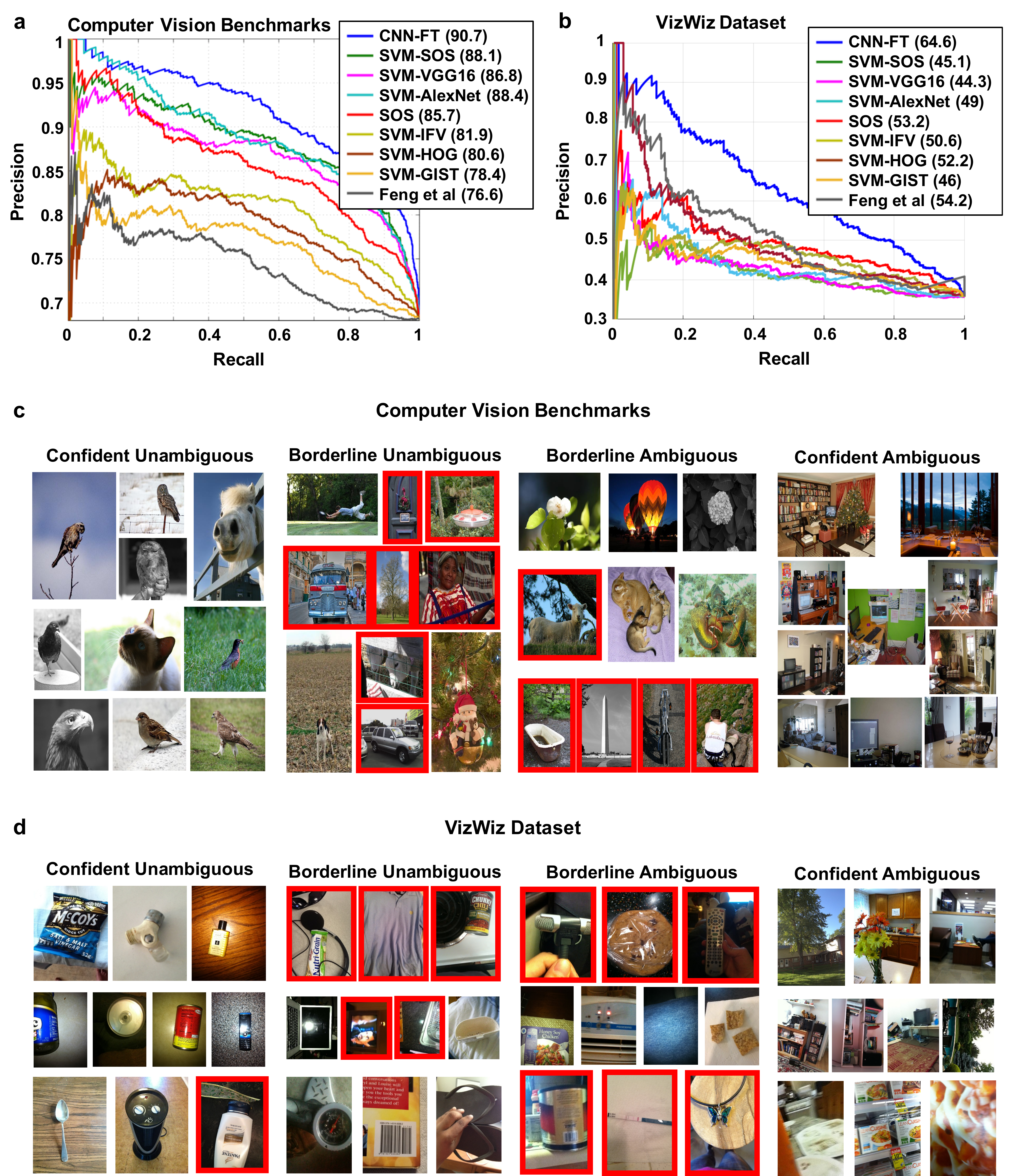}
\caption{(\textbf{a, b}) Precision-recall curves and average precision scores for nine benchmark systems tested on images coming from (\textbf{a}) seven computer vision benchmarks and (\textbf{b}) blind photographers.  Our \textit{STATIC} fine-tuned CNN classifier (\texttt{CNN-FT}) outperforms SVMs trained on off-the-shelf CNN features (AlexNet, VGG16, SOS) and gradient-based features (GIST, HOG, IFV) as well as existing saliency (Feng et al~\cite{FengWeTaZhSu11}) and subitizing (SOS, Zhang et al~\cite{ZhangMaSaScBeLiShPrMe15}) methods.  (\textbf{c, d}) Also shown are example foreground object ambiguity predictions from our deep learning system, \texttt{CNN-FT}, for the images coming from (\textbf{c}) seven computer vision benchmarks and (\textbf{d}) blind photographers.  We show the top ten images most confidently predicted as ambiguous/unambiguous and borderline cases.  See images with red boundaries for examples where the classifier predicts the wrong label.  Images with a single, unambiguous object are successfully found despite the large diversity in the object's appearance (e.g., shape, color, texture).  }
\label{fig_precisionRecallCurve}
\vspace{2em}
\end{figure}

\vspace{-0.5em}\paragraph{Predictive Performance.}

\textbf{Figures~\ref{fig_precisionRecallCurve}a,b} show the precision-recall curves for all models.  Our network fine-tuned for \textit{STATIC}, \texttt{CNN-FT}, achieves the best overall performance.  For example, the average precision (AP) score improves over the top-performing baseline by five percentage points (i.e., 85.7\% for \texttt{SOS} baseline versus 90.7\% for \texttt{CNN-FT}) and over 10 percentage points (i.e., 54.2\% for \texttt{Feng et al.} baseline~\cite{FengWeTaZhSu11} versus 64.6\% for \texttt{CNN-FT}) for the computer vision benchmarks and VizWiz images respectively.  Our results confirm it is possible to predict whether a novel image contains a single, unambiguous foreground object.  This is interesting because it indicates that image content alone---without external psychological cues---often carries sufficient information to gauge ambiguity.

Overall, we observe classifiers perform worse on the VizWiz images than the computer vision benchmarks; e.g., AP scores for the top-performing \texttt{CNN-FT} classifiers are 90.7\% for computer vision benchmarks and 64.6\% for VizWiz images.  While our findings demonstrate the promise of automating the proposed prediction task, they also reveal an important, largely unsolved challenge for modern computer vision tools in handling poor quality images commonly captured by blind photographers.

%Overall, we also found that SVMs trained with CNN features (\texttt{SVM-SOS}, \texttt{SVM-VGG16}, \texttt{SVM-AlexNet}) typically performed better on the computer vision benchmarks and worse on the VizWiz images than SVMs trained with gradient-based features (\texttt{SVM-GIST}, \texttt{SVM-HOG}, \texttt{SVM-IFV}).  While deep learning features often yield greater discriminative power than handcrafted features, our findings suggest CNN-based features may be over-fitting to high quality images curated from the web.  There is a domain shift from ImageNet to VizWiz, highlighting the value of further work to improve the transferability of such features to new domains.%Our finding parallels other reported findings that deep learning systems are vulnerable to challenging/adversarial examples in real world scenarios~\cite{KurakinGoBe16}.

\textbf{Figures~\ref{fig_precisionRecallCurve}c,d} show prediction results from \texttt{CNN-FT}.  Specifically, each figure shows 10 images for four categories: confident (un)ambiguous and borderline (un)ambiguous.  The borderline images highlight that the predictor often is confused by images with semi-dominant objects neighboring distractor objects.  The images with the most confident ``unambiguous" predictions highlight that the predictor expects human agreement in the presence of a dominant object against a consistently textured background.  Interestingly, the predictor does not appear to make strong assumptions about the appearance of the foreground object, as exemplified by the dominant objects exhibiting various shapes, sizes, colors, and textures as well as being positioned in various parts of the images.

\begin{figure*}[t!] 
\centering
\includegraphics[width=1\textwidth]{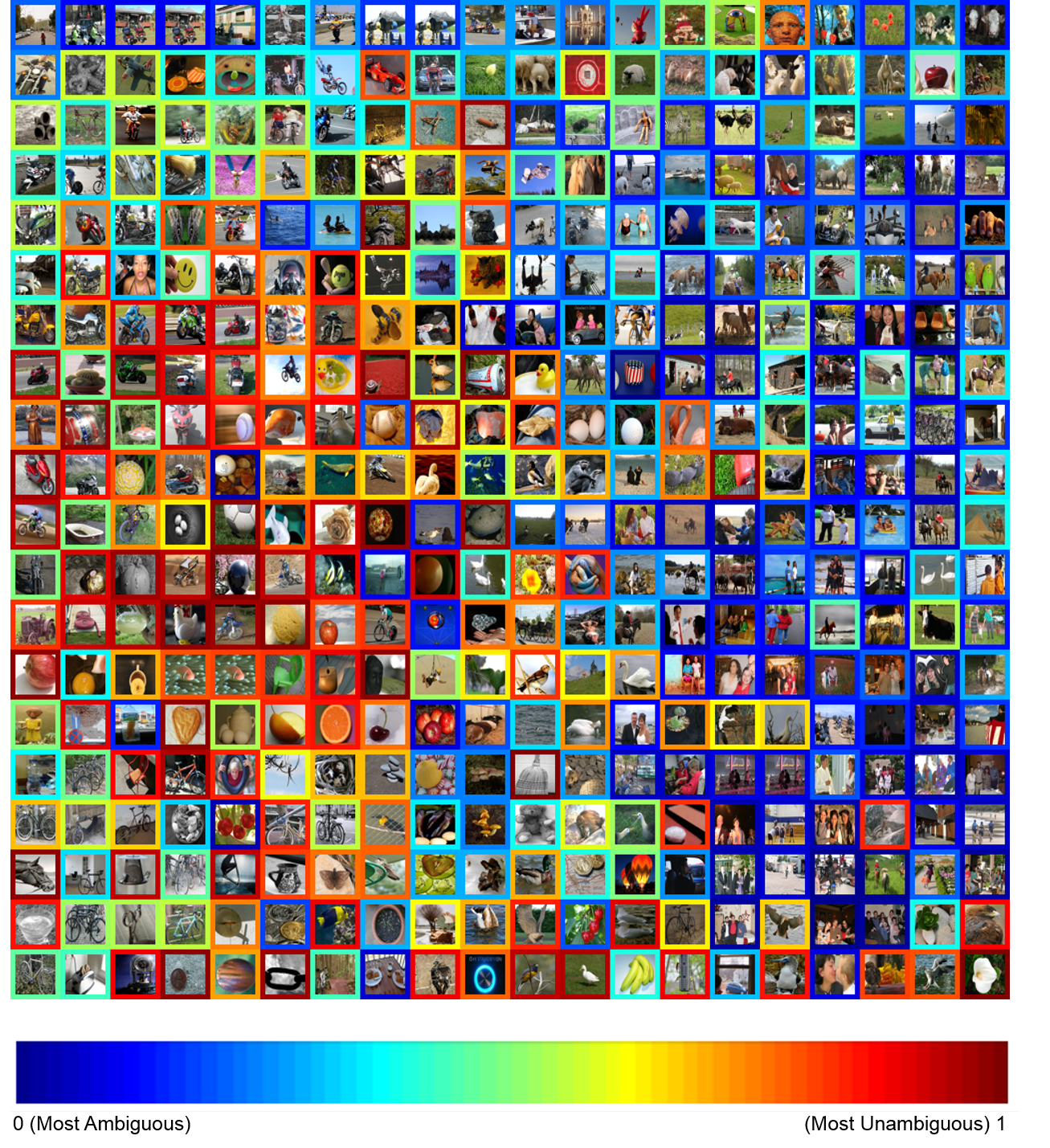}
\caption{Images neighbor other images that share similar values in the final 4096-dimensional vector layer of the CNN.  The border color for each image indicates the classification score, with blue (0) reflecting most confident ambiguous and red (1) reflecting most confident unambiguous.  (Best viewed in color).}
\label{fig_tSNEVisualization}
\end{figure*}

\vspace{-0.5em}\paragraph{Predictive Cues.}
We apply t-SNE, a visualization technique, to the seven computer vision benchmarks in the STATIC test dataset in order to offer further insight into what our top-performing fine-tuned \texttt{CNN-FT} classifier learned.  We leverage publicly-available code\footnote{\tt http://cs.stanford.edu/people/karpathy/cnnembed/} to create the visualization (\textbf{Fig.~\ref{fig_tSNEVisualization}}).  This 2D t-SNE plot places images close together that have similar learned descriptors in our CNN that is fine-tuned to target the (un)ambiguity label.  Based on observed image clusters, we posit the system is picking up on a combination of low-level visual features.    

First, we observe the images in the bottom left quadrant tend to capture visually similar circular objects.  For example, often neighboring images share similar colored circular objects (e.g., multi-colored rock and disc in the bottom row; orange/red circular fruit in the 5th from bottom row). In addition, some image clusters also show circular objects with similar diameters (e.g., bicycle rim and water glass rim in bottom row).  While the specific objects visible may vary, the features have picked up on generic properties that can lead to (un)ambiguity.

%Second, we observe the images in the right half tend to show objects with multiple vertical structures in complex images.  For example, the bottom right quadrant tends to show images with numerous people or animals.  Moving upwards, the type of vertical structures often are found in animals such as long legs of horses and camels, long necks of swans, and long bodies of birds.  

We also observe in the top left quadrant that images tend to show bikes.  However, in that cluster, the classifier seems to have picked up on generic properties for separating the object category based on (un)ambiguity; i.e., the upper half is ambiguous (images with blue boundaries) and lower half is unambiguous (images with red boundaries).  This suggests the classifier is learning ambiguity-specific properties for separating images rather than following object category lines.  We speculate this observation explains how, in \textbf{Figure~\ref{fig_motivation}b}, two visually similar images can lead to different labels.  Specifically, humans perceive an image showing a single flower as \emph{unambiguous} and an image showing a collection of crayons shaped like a flower as \emph{ambiguous}.  Our classifier seems to similarly be leveraging ambiguity-specific properties to detangle visually similar content and decide (un)ambiguity.  

%Specifically, humans perceive an image showing a single flower as \emph{unambiguous} and an image showing a collection of crayons shaped like the petals of a flower as \emph{ambiguous}.  We hypothesize the different labels arise as a result of whether the individual parts are recognized as meaningful objects on their own versus meaningful parts of another object -- i.e., crayons are meaningful in the absence of flowers while petals are not.  Our classifier may similarly be leveraging part detections (e.g., petals) with the number of plausible object detections (e.g., flower, crayon) to decide (un)ambiguity.  

%----------------------------------------------------------------------------------------
\section{How Many Object Segmentations to Solicit?}
\label{sec_crowdsourcingSegProblem}
%----------------------------------------------------------------------------------------
As observed in \textbf{Section~\ref{sec:static}}, the problem of foreground object ambiguity is of immediate practical relevance for evaluating algorithms on existing object-centric datasets (\textbf{Table~\ref{table_crowdPerceptionStudy}}).  In particular, we currently we lack benchmarks that include the diversity of valid foreground object segmentations for a batch of images.  In this section, we propose a system to efficiently create such benchmarks.  Today's status quo is to ask $N$ \emph{independent viewers} to locate the single most prominent object in a given image~\cite{AchantaHeEsSu09,AlpertGaBaBr07,BorjiSiIt13,ChenMiHuToHu15,JiangWaYuWuZhLi13,LiuYuSuWaZhTaSh11}\footnote{Collecting annotations from multiple \emph{independent} annotators is necessary to avoid annotator bias (e.g., Berkeley Segmentation Dataset~\cite{MartinFoTaMa01}, MSRA~\cite{LiuYuSuWaZhTaSh11}).  As discussed in Section~\ref{sec_RelatedWork}, this is in stark contrast to dataset collection systems that solicit redundant annotations by showing each new annotator all previously-collected segmentations overlaid on the image (e.g., LabelMe~\cite{RussellToMuFr08}, VOC~\cite{EveringhamGoWiWiZi10}, MSCOCO~\cite{LinMaBeHaPeRaDoZi14}).  This design difference stems from different aims.  While the latter aims to annotate all objects (possibly only for a pre-defined set of object categories) in an image, the former focuses on localizing all objects deemed the \emph{single most prominent object} according to human perception.}.  Commonly, a uniform number of annotations are collected for every image, ranging from as few as one segmentation per image~\cite{ChenMiHuToHu15} to as many as ten (bounding box) annotations per image~\cite{LiuYuSuWaZhTaSh11}.  Our aim is to capture the diversity of valid ground truths across all images without uniformly segmenting each image $N$ times.  

Our method is related but distinct from the Welinder and Perona~\cite{WelinderPe10} method, which also dynamically decides the number of redundant object segmentations to collect per image.  However, as discussed in Section~\ref{sec_RelatedWork}, our goal is very different.  While prior work aims to efficiently achieve a desired level of confidence in a \emph{single ground truth} per image~\cite{WelinderPe10,WhitehillWuBeMoRu09,WelinderBrBePe10}, our system is designed to efficiently capture \emph{annotation diversity} and so all valid ground truths per image.  In fact, our method fills a gap in the literature the authors themselves report---errors for their method are concentrated on cases where ``intrinsic uncertainty of the ground truth label is high"~\cite{WelinderPe10}.  Our system decides the number of human annotators to recruit based on whether the image is deemed unambiguous (single) versus ambiguous (multiple) by our \textit{STATIC} ambiguity predictor.

Our system begins with exactly one human-drawn foreground object segmentation for each image.  Given bounded annotation resources, the system can only request additional annotations for a subset of the images.  Our goal is to capture as much of the diversity of valid foreground object segmentations as possible for the batch with the allocated human annotation budget.  Our key design decisions are how to 1) allocate annotation effort and 2) quantify diversity captured by human-drawn foreground object segmentations.

%Collecting and fusing redundant segmentations is a widely-adopted method~\cite{AlpertGaBaBr07,MartinFoTaMa01,WarfieldZoWe04}.  
%To our knowledge, our redundancy allocation system is the first to efficiently capture \emph{annotation diversity}; prior work solicits redundant labels to efficiently achieve a desired level of confidence in a \emph{single ground truth} per image~\cite{SheshadriLease13,WelinderPe10,WhitehillWuBeMoRu09,WelinderBrBePe10}.  As discussed in Section~\ref{sec_RelatedWork}, these are two very different goals, even though both entail deciding how many annotations to obtain for an image.  

\vspace{-0.5em}\paragraph{Allocating Human Annotation Effort.}  
Our system takes in a batch of $N$ images with a redundancy budget $B$ indicating the number of images to receive redundant human annotations.  The system first collects one segmentation for every image.  Then, the system applies our proposed prediction system discussed in Section~\ref{sec:predict} to every image in the batch; we use our \texttt{CNN-FT} method trained on the computer vision benchmarks.  Next, the system orders the $N$ images based on predicted scores from the classifier, from most confidently predicted ``ambiguous" images to the most confidently ``unambiguous" images.  Finally, the system greedily assigns the given budget of annotation effort for redundancy to the $R$ images predicted to reflect the greatest likelihood of ambiguity.  Each image assigned to receive redundant labels is allocated a fixed number of additional human annotations.  

\vspace{-0.5em}\paragraph{Measuring Segmentation Diversity.}  
We now describe our method to evaluate the segmentation diversity captured by the collection of human-drawn segmentations for a batch of images $I = \{1,...,N\}$.  Given the subset of images $R$ assigned to receive $A$ redundant annotations each, we compute total diversity as follows:

\vspace{-1.5em}
\begin{eqnarray}
%D(I) = \sum_{k=1}^{N} d_{k0} + \sum_{j=1}^{B} \sum_{a=1}^A d_{ja}
D(I) = \sum_{k \in I} d_{k0} + \sum_{j \in R} \sum_{a=1}^A d_{ja} 
\label{eqn_diversityScore}
\end{eqnarray}

\noindent
where $d_{k0}$ represents the diversity captured by the first annotation for the $k$-th image (defined below), $d_{ja}$ represents the diversity captured by the $a$-th redundant annotation for the $j$-th image, and $D(I)$ reflects the total annotation diversity captured for image batch $I$.  The first term evaluates the diversity captured by a single segmentation per image.  With no redundancy budget, the total diversity will come from this term.  The second term evaluates the diversity captured by redundant annotations.  When the maximum redundancy budget is available, total diversity will include the diversity captured by having redundant annotations for every image.  Given a partial redundancy budget, if the ambiguity predictions were perfect, then we could safely solicit a single human-drawn segmentation on unambiguous images and extra segmentations for ambiguous images.

\begin{figure}[t!]
\centering
\includegraphics[width=0.88\textwidth]{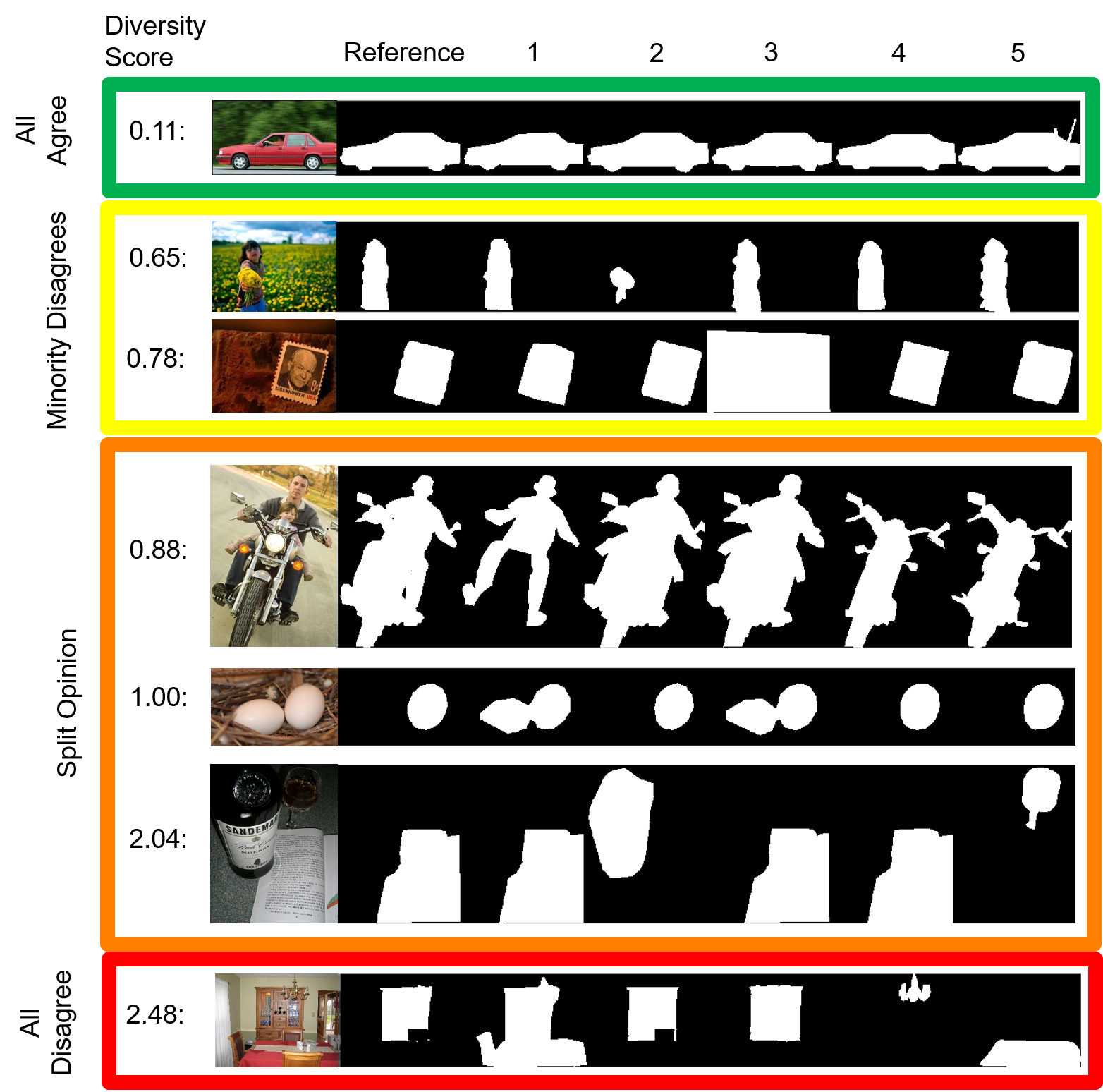}
\caption{Illustration of the measured segmentation diversity among five crowdsourced segmentations.  Each example shows the raw image, diversity score derived with the weighted F-measure~\cite{MargolinZeTa14}, reference, and five crowdsourced segmentations.  These illustrate how the following different levels of disagreement correlate to computed diversity scores: All Agree, Minority Disagrees, Split Opinions, and All Disagree.  Diversity is often least when ``All Agree" and greatest when ``All Disagree".}
\label{fig_diversityTypes}
\end{figure}

Our goal is to choose diversity measures that reveal when humans disagree because of ambiguity versus minute differences in boundary detail (e.g., \textbf{Figure~\ref{fig_diversityTypes}}, rows 2-7 versus row 1).  We chose two diversity measures that indicate for an image how different each \emph{individual}'s annotation is from the \emph{reference} segmentation.  The reference segmentation represents the pixel majority vote result from multiple annotators' segmentations.  Diversity is measured as the difference of a human-drawn segmentation $S$ to the reference segmentation $R$.  One measure is region-based and is computed by 1 - $sim(S, R)$, where $sim$ is the weighted F-measure~\cite{MargolinZeTa14}.  This measure computes the number of pixels in common between the two segmentations, using both the dependency between neighboring pixels and the location of the errors.  The second measure is boundary-based and, in particular, we compute the Chamfer distance between $S$ and $R$, which indicates the distance between two shapes.  For both measures, larger values reflect greater diversity.

%As shown in row 1 of \textbf{Figure~\ref{fig_diversityTypes}}, when all human annotators segment the same object (e.g., car), there is very little diversity and so a low value.  On the other hand, as shown in row 7, when all human annotators segment different objects in the image, there is a lot of diversity and so a large value.  Rows 2-6 of \textbf{Figure~\ref{fig_diversityTypes}}, illustrate when multiple human annotators segment the same object and one to a few annotators segment a different object.  In this case, diversity values typically fall between that observed with no agreement and total agreement.

\vspace{-0.5em}\paragraph{Experimental Design.}
We evaluate the impact of selectively allocating human effort to create foreground object segmentations as a function of the available budget of human effort.  For each budget level, we measure the total diversity resulting for the batch of images.  When a system does well, the segmentation diversity captured will remain high despite using a lower budget.

\begin{figure*}[t!] 
\centering
\includegraphics[width=1\textwidth]{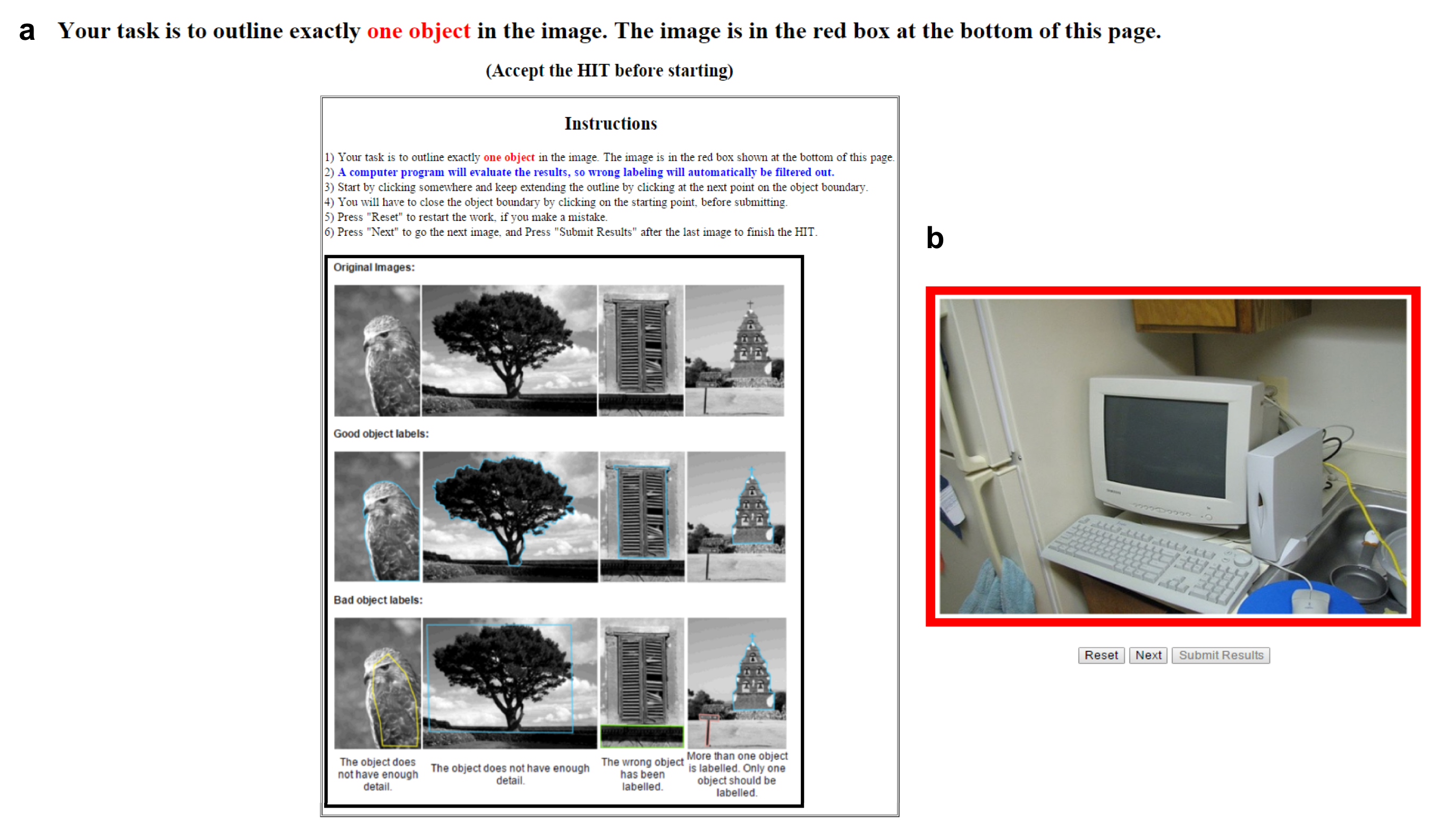}
\caption{Foreground object segmentation system \textbf{(a)} instructions and \textbf{(b)} user interface.}
\label{fig_crowdSegmentationInstructionsAndUI}
\end{figure*} 

We conduct our studies on 800 randomly selected images from the STATIC test set.  We collect segmentations from crowd workers recruited from Amazon Mechanical Turk (AMT).  We limit our pool of workers to those who previously completed at least 100 tasks and received at least a 92\% approval rating.  Our task includes instructions at the top of the webpage (\textbf{Fig.~\ref{fig_crowdSegmentationInstructionsAndUI}a}) followed by the image to segment at the bottom of the webpage (\textbf{Fig.~\ref{fig_crowdSegmentationInstructionsAndUI}b}).  The user interface restricts the worker to exactly one object segmentation per image.  We include five images per HIT and pay workers \$0.10 to complete each HIT.

We compare our system to the following baselines:

\begin{description}
\setlength\itemsep{-0.1em}
\item [\texttt{W\&P-BB\cite{WelinderPe10}}]: An online crowdsourcing system from Welinder and Perona~\cite{WelinderPe10} which decides the redundancy level per image for all images.  Specifically, given a threshold, annotations are collected for an image until annotation agreement exceeds the threshold.  Agreement is measured using both a confidence in the annotators' skills and bounding box similarity.  We sweep through all thresholds to create a human effort budget versus diversity curve.  We use the bounding boxes of our crowdsourced segmentations.  
 %We use the authors' publicly available code\footnote{{\tt https://github.com/welinder/cubam/blob/public/cubam/BinaryBiasModel.py}} to compute agreement, using as input the bounding boxes of the crowdsourced segmentations.  
\item [\texttt{W\&P-Seg\cite{WelinderPe10}}]: A system matching \texttt{W\&P-BB} except that our crowdsourced segmentations are used directly to measure annotation agreement.  
\item [\texttt{SOS\cite{ZhangMaSaScBeLiShPrMe15}}]: A method that predicts a confidence in whether the image contains 0, 1, 2, 3, or 4+ salient objects.  Images are ordered by most confident to least confident predictions for 1 object followed by least to most confident predictions for 2, 3, 4+, and 0 objects respectively.  Images ranked least confident are prioritized to receive redundancy.  
\item [\texttt{Status Quo}]: Images are randomly prioritized to receive redundancy.
\item [\texttt{Perfect}]: Images are ordered by total diversity score per image, based on having \emph{all} segmentations.  This demonstrates the best a system could achieve.
\end{description}

% With our tool, workers trace the boundary of an object by clicking points on the image which are connected with straight lines.  A worker completes the segmentation by clicking on the first clicked point.  The worker can at any time click ``Reset" to delete the current annotation and redraw the object.  

To evaluate our approach using existing salient object detection redundancy levels, we investigate performance for two redundancy levels.  First, we evaluate performance by employing the commonly-employed redundancy level of \emph{five} annotations per image (e.g., MSRA-A~\cite{LiuYuSuWaZhTaSh11}, ~\cite{AlpertGaBaBr07,MartinFoTaMa01,WarfieldZoWe04}).  We also employ the more rigorous redundancy level of \emph{ten} annotations per image (i.e., MSRA-B~\cite{LiuYuSuWaZhTaSh11}).

\begin{figure}[b!]
\centering
\includegraphics[width=0.94\textwidth]{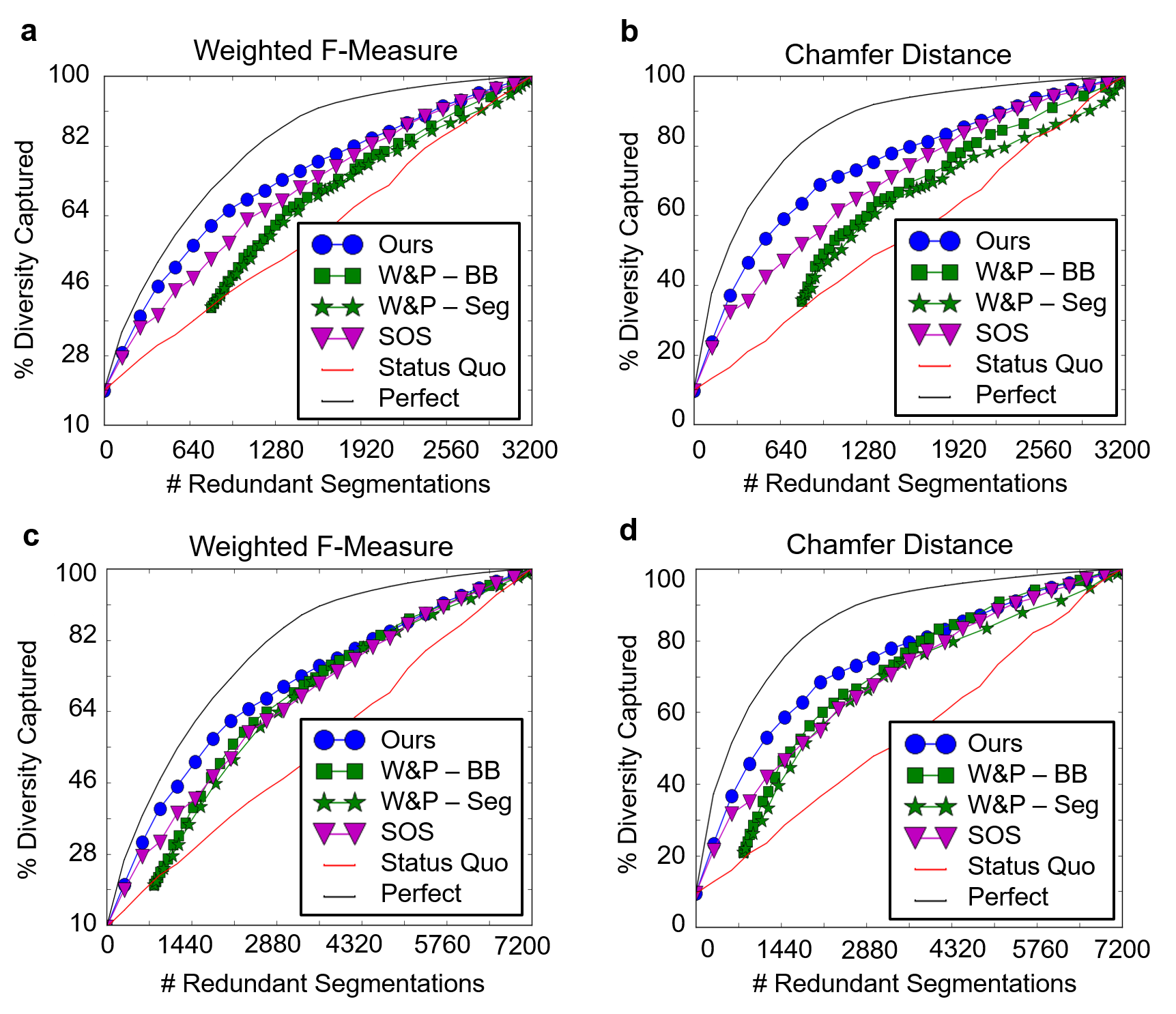}
\caption{(\textbf{a, b, c, d}) The plots show the value of our prediction system and five baselines to allocate redundant segmentations for capturing foreground object diversity for 800 images.  Results are shown based on two diversity measures: Weighted F-Measure and Chamfer Distance.  Boundary conditions are (leftmost) ``no redundancy per image" for all images and (rightmost) (\textbf{a, b}) ``four redundant segmentations per image" or (\textbf{c, d}) ``nine redundant segmentations per image" for all images.  Our system outperforms existing online crowdsourcing systems (\texttt{W\&P}~\cite{WelinderPe10}) and a saliency (SOS, Zhang et al~\cite{ZhangMaSaScBeLiShPrMe15}) method.}
\label{fig_redundancyVsDiversityPlot}
\end{figure}

\vspace{-0.5em}\paragraph{Experimental Results.}

Our approach consistently outperforms the baselines with respect to both diversity measures for both the redundancy level of five segmentations per image (\textbf{Fig.~\ref{fig_redundancyVsDiversityPlot}a, b}) and ten segmentations per image  (\textbf{Fig.~\ref{fig_redundancyVsDiversityPlot}c, d}).  For example, our system accelerates the collection of $\sim$51\% of the diversity by at least 27\% over the four baselines (\texttt{W\&P-BB}, \texttt{W\&P-Seg}, \texttt{SOS}, \texttt{Status Quo}), with respect to the region-based measure (\textbf{Fig.~\ref{fig_redundancyVsDiversityPlot}d}).  In absolute terms, this translates to eliminating over six human-hours of annotation time for 800 images, assuming a human takes approximately 54 seconds to segment an object~\cite{JainGr13}.  In addition, with respect to the boundary-based measure, our system accelerates the collection of $\sim$53\% of the diversity by as much as 47\% over the four baselines (\textbf{Fig.~\ref{fig_redundancyVsDiversityPlot}d}).  In absolute terms, this means eliminating over eight human-hours of annotation time for 800 images.  Our performance gains taper for both measures when our system has captured most of the total diversity ($\sim$70\%).  Our findings offer promising evidence that it is possible to \emph{efficiently} address the issue of image ambiguity and its effect on foreground object segmentation evaluation (discussed in Section~\ref{sec_ambiguityAndSegEval}) by collecting extra annotations only for images where a diversity of ground truths are expected.

We attribute our advantage over the top-performing, saliency-based predictor (i.e., \texttt{SOS}) to the observation that ambiguity arises for a variety of causes beyond multiple salient objects, including object granularity and occlusion (\textbf{Fig.~\ref{fig_redundancyVsDiversityPlot}c}).  Our findings highlight a value in directly predicting whether humans will agree on a single foreground object rather than predicting the number of detected salient objects in an image.   

Our method also significantly outperforms methods that predict the exact number of annotations to collect per image, i.e., the \texttt{W\&P} baselines\footnote{This method requires a minimum of two segmentations per image and allocates a different number of additional annotations for different images.}.  We attribute our advantage to the fact that the \texttt{W\&P} baselines only solicit additional segmentations \emph{if} annotator disagreement is already observed between the first two segmentations.  Yet, as shown in \textbf{Figure~\ref{fig_diversityTypes}}, one may need to collect more than two foreground object segmentations to observe diversity.  Our findings highlight a value in directly predicting from an image whether humans will agree on a single foreground object rather than making inferences from observed human annotations.   

Finally, as observed in \textbf{Figure~\ref{fig_diversityTypes}}, a different number of valid interpretations can arise due to ambiguity (e.g., 2, 3, 4, or 5 valid ground truths).  This observation motivates two valuable areas for future work to achieve further savings: 1) predict the exact number of foreground object segmentations in an image and 2) target the images to the appropriate crowd workers predicted to, as a group, produce the diversity of valid outcomes without duplicates.

% Crowd = 274/719 = 38% of diversity
% baseline = 668
% us = 268
% (668-268)/268*100 = 1.27

% Crowd = 442/719 = 61% of diversity
% us = 800
% random baseline = 1732
% (1732-800)/800*100 = 1.16

% W&P-BB baseline = 1278
% (1278-800)/800*100 = 60\%

%%--------------------------------------------------------------------------------
\section{Conclusions}
%%--------------------------------------------------------------------------------
Our work reveals a promising, largely-untapped research problem of accounting for foreground object ambiguity to improve computer vision systems.  We established a benchmark and showed segmentation algorithms are getting penalized when they produce valid results.  We proposed a system that accurately predicts whether an image is ambiguous and so should have multiple ground truths, both for images in established vision benchmarks and from blind photographers.  Finally, we demonstrated how to reduce human effort to collect the diversity of valid foreground object segmentations for a batch of images, improving upon existing saliency-based and online crowdsourcing methods. 

We offer this work as a valuable step towards a larger community effort to build modern vision systems that account for the foreground object ambiguity that humans perceive.  One future research direction includes human computer interaction studies to explore how blind photographers prefer to integrate foreground ambiguity predictions with existing mobile phone camera applications to help them recognize objects in their environment.  Future work also includes exploring how to more efficiently create salient foreground object benchmarks that include the diversity of foreground object segmentations.  

%Possible future research directions also include running the study on a larger image set and quantitatively analyzing the causes in images that influence the successes and failures of the different algorithms and initial contours.

% to more efficiently utilize crowdsourcing by analyzing the reliability of crowdsourced workers and what number of annotations are necessary.

%Another possible future research direction  will investigate predicting the exact number of most prominent foreground objects that exist in an image to achieve further cost savings when crowdsourcing.

%Modern vision systems currently make mistakes because they do not account for the foreground object ambiguity [1, 2, 8, 11, 12, 33] that we know humans perceive [18, 27].

%Future work will investigate whether fusion methods are preferred for different applications and different annotator levels such as non-experts as well as whether more sophisticated fusion method formulations are preferred.

%%--------------------------------------------------------------------------------
\section*{Acknowledgments}
%%--------------------------------------------------------------------------------
\noindent
The authors gratefully acknowledge funding from the Office of Naval Research (ONR YIP N00014-12-1-0754) and National Science Foundation (IIS-1421943) and thank the anonymous crowd workers for participating in our experiments.  

{\small
\bibliographystyle{ieee}
\bibliography{myReferences}
}

\end{document}